\documentclass[letterpaper]{article} 
\usepackage{aaai2027}  
\usepackage[hyphens]{url}  
\usepackage{graphicx} 
\usepackage{natbib}  
\usepackage{caption} 
\usepackage{algorithm}
\usepackage{algorithmic}

\usepackage{newfloat}
\usepackage{listings}
\DeclareCaptionStyle{ruled}{labelfont=normalfont,labelsep=colon,strut=off} 
\floatstyle{ruled}
\newfloat{listing}{tb}{lst}{}
\floatname{listing}{Listing}

\usepackage{times} 
\usepackage{latexsym}

\usepackage[T1]{fontenc}

\usepackage[utf8]{inputenc}

\usepackage{microtype}

\usepackage{tcolorbox}
\usepackage{pifont} 
\usepackage{adjustbox} 
\usepackage{booktabs} 
\usepackage{amsmath} 

\usepackage{xcolor, colortbl}
\usepackage{listings}
\usepackage{makecell}
\usepackage{multirow}
\usepackage{array}
\usepackage{caption}
\usepackage{tabularx}
\usepackage{inconsolata}

\usepackage{graphicx}
\usepackage{newfloat}
\usepackage{pifont}
\usepackage{amsmath} 
\usepackage{makecell}
\usepackage{booktabs}
\usepackage{amssymb}
\usepackage{enumitem}
\usepackage{pifont}
\usepackage{subcaption}
\usepackage{placeins}
\usepackage{bm}
\newcommand{\cmark}{\ding{51}} 
\newcommand{\xmark}{\ding{55}} 
\definecolor{BrightGreen}{RGB}{0,120,0}

\nocopyright 

\title{Enhancing Localized Reasoning for Long Video Understanding via Efficient Segment-to-Video Supervision}

\author{
  Beibei Zhang$^{1,2}$,
  Chao Xu$^{2}$, 
  Jun Lan$^{2, ^\ddagger}$, 
  Zongyi Li$^{2}$, 
  Lai Wei$^{3}$, 
  Huijia Zhu$^{2}$, 
  Tongwei Ren$^{1,^* }$
}
\affiliations{
  $^1$State Key Laboratory for Novel Software Technology, Nanjing University \quad $^2$ Ant Group \\
  $^3$School of Computer Science, Shanghai Jiao Tong University

  zhangbb@smail.nju.edu.cn, \{yanyue.xc, yelan.lj, lizongyi.lzy, huijia.zhj\}@antgroup.com, waltonfuture@sjtu.edu.cn, rentw@nju.edu.cn

}

\begin{document}

\maketitle

\begingroup
\renewcommand{\thefootnote}{}
\footnotetext{\footnotesize
$^\ddagger$ Project Lead.
$^*$ Corresponding Author.
This work was done during the first author's internship at Ant Group.
}
\addtocounter{footnote}{-1}
\endgroup

\begin{abstract}
Though Multimodal Large Language Models (MLLMs) have shown impressive potential in video understanding, 
long video understanding (LVU) remains challenging since distracting noise in complex and lengthy contexts can obscure localized details, misleading MLLMs to produce incorrect answers.
Recent works mitigate these issues by incentivizing deep reasoning to include relevant evidence.
However, these methods have two main problems: 
First, the reinforcement fine-tuning framework (RFT) they leveraged incurs substantial training overheads, including high annotation costs and complicated reward designs.
Second, the self-reflective and iterative-perception mechanism in some methods causes lengthy outputs and high inference latency.
To alleviate these problems, we propose a novel \emph{\textbf{S}egment-\textbf{t}o-\textbf{V}ideo Supervision} method (S2V) to efficiently enhance fine-grained reasoning in LVU.
Specifically, we generate question answer pairs (VQA) based on localized segments, and then transfer these segment-based VQA back to the whole video for training. 
Due to focusing on short segments, segment-based VQA can naturally notice details which tend to be overlooked from a whole-video perspective.
Training on such data can enforce MLLMs to correctly associate fine-grained details with QA while avoiding distracting noise in the whole video.
The S2V training involves just reinforcement learning (RL) with a simple accuracy reward based on only 10K VQA samples and the resulting S2V model predicts answer using a single forward pass with limited output tokens.
Experimental results demonstrate that S2V can consistently improve LVU performance across multiple LVU benchmarks, outperforming both general MLLMs and reasoning-based methods not only in LVU accuracy but also in training and inference efficiency. 

\end{abstract}
 
\begin{figure}[!t]        
  \centering        
  \includegraphics[width=1\columnwidth]{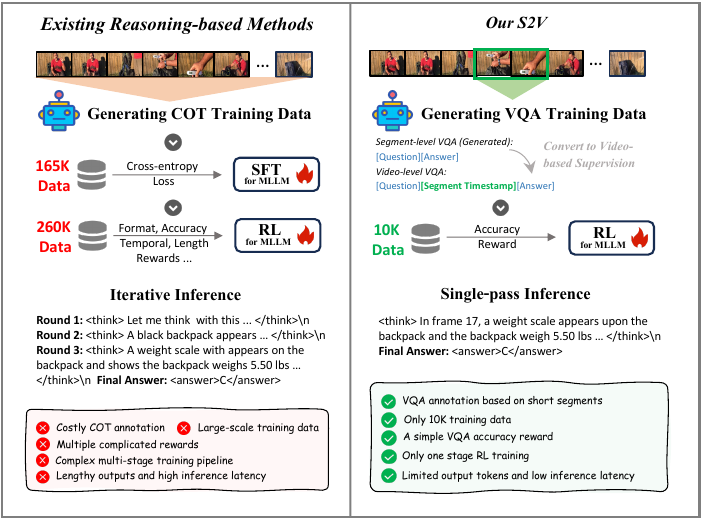}     
  \caption{Comparison between existing reasoning-based LVU methods \emph{vs.} our S2V.}
\label{fig:motivation} 
\end{figure} 
   
\section{Introduction} 
Recent advancements in MLLMs have demonstrated remarkable success in video understanding~\cite{gpt5, qwen3vl, internvl35,minicpm45v}.
Nevertheless, large amounts of distracting noise in long-form context still hinder MLLMs from discriminating and comprehending useful details for LVU.
Previous solutions rely on an additional retrieval model to identify query-relevant evidence, which are then fed into MLLM for answer generation~\cite{liu2025bolt, yao2025gens, kim2025salova}. 
However, these retrieval models inevitably incur extra training and inference overheads.

\begin{figure*}[!t]             
  \centering        
  \includegraphics[width=1\textwidth]{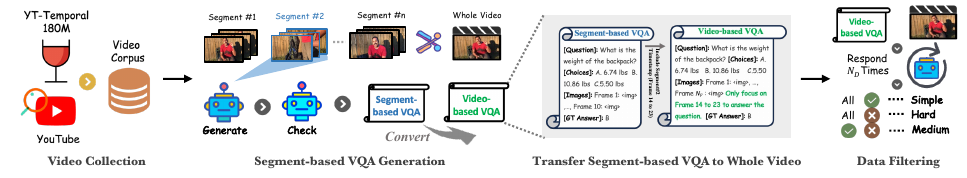}    
  \caption{Overview of the S2V data synthesis pipeline. Here, $n$ refers to the segment number, which is a dynamic number depending on the segment detection result, $N_F$ is the sampled frame number of the whole video and $N_D$ is the response repetition number for difficulty filtering.}
\label{fig:framework}    
\end{figure*}  

Inspired by the success of reasoning-oriented models such as DeepSeek-R1~\cite{guo2025deepseek}, recent works shift focus toward incorporating query-relevant evidence in MLLM reasoning~\cite{feng2025videor1, wang2025videorft, qiu2026longvideor1}.
These methods typically leverage a RFT framework for multi-task optimization, \emph{e.g.,} evidence grounding and question answering.
The RFT framework includes two-stage training: supervised fine-tuning (SFT) for cold-start initialization and reinforcement learning (RL) for model refinement.
To adapt to various output formats and acquire basic reasoning capabilities, the SFT phase requires large-scale COT annotations, which are high-cost for using expensive annotation models like Gemini~\cite{gemini25} to generate long COTs.
And the RL stage relies on complicated reward designs tailored for each task to improve reasoning accuracy.
For example, Video-R1~\cite{feng2025videor1} constructs 165K COT data for SFT and 260K RL data and designs four different rewards for RL optimization.
Moreover, to collect more necessary evidence, some methods~\cite{hu2026spectemp, yan2025videochatr15, wang2025videorts} include the self-reflective and iterative-perception mechanism in the reasoning process, where the long reasoning outputs, e.g., more than 1k text tokens, lead to high inference latency.
This also becomes a crucial bottleneck in real-world video applications.

To alleviate these problems, we propose a novel and easy-to-implement \emph{Segment-to-Video Supervision} method to efficiently enhance fine-grained reasoning for LVU, significantly improving the LVU accuracy with limited training and inference costs.
Long videos are composed of multiple relatively-independent yet semantically-interconnected segments.
These segments support LVU questions and are often treated as search units in LVU reasoning~\cite{kim2025salova, yan2025videochatr15, hu2026spectemp}.
We firstly generate question-answer pairs based on short segments.
Then, by including segment timestamp in task instruction, we transfer these segment-based VQAs back to the whole video, resulting in unambiguous video-based VQAs.
By focusing on short segments, segment-based annotations can naturally involve fine-grained details which tend to be overlooked from a whole-video perspective.
Training on such data can encourage MLLMs to associate fine-grained details with QA and suppress distracting noise in the whole video.

Existing evidence-reasoning datasets mostly annotate evidence from the whole video, guided by questions generated from whole-video perspective~\cite{hu2026spectemp, jiang2026videop2r, meng2025openo3}.
This results in a lack of VQA samples that primarily focus on short segments, limiting the ability to comprehend localized fine-grained details.
To fill this gap and support S2V training, we propose a fully automated pipeline to construct robust segment-to-video supervision data, resulting in a S2V-10K dataset.

Compared with methods driven by large-scale training data, S2V significantly reduces training costs by using only 10K samples.
Meanwhile, S2V generates only VQA annotations based on short segments, whereas other methods construct long CoT annotations from whole videos, thereby substantially reducing annotation complexity and expenses.
Furthermore, S2V significantly simplifies the training pipeline by using only RL with a simple VQA accuracy reward.
And S2V predicts answers through a single MLLM forward pass with limited output tokens, achieving lower inference latency than iterative inference methods.

We conduct extensive experiments to validate the effectiveness of S2V.
Ablation studies show that S2V can better utilize fine-grained evidence in complex long-video contexts, consistently improving LVU performance over base MLLMs.
Moreover, experimental results across multiple LVU benchmarks~\cite{fu2025videomme,zhou2024mlvu,wu2024longvideobench} demonstrate our S2V can also surpass other general MLLMs and LVU-tailored methods, including both retrieval-based and reasoning-based methods, not only in accuracy but also in training and inference efficiency.
In general, the main contributions of this work can be summarized as follows:
 
\begin{itemize}[leftmargin=*, itemsep=0pt, topsep=2pt, parsep=0pt, partopsep=0pt]
\item We propose a novel \emph{Segment to Video Supervision} method to efficiently enhance fine-grained reasoning for LVU. 
This method enforces MLLM to comprehend localized details from distracting long contexts by transferring segment-based annotations to the whole video for training.
\item We design a fully automated pipeline to synthesize robust segment to video supervision data and contribute a S2V-10K dataset to support S2V training. 
\item Compared to existing reasoning-based LVU methods, the proposed S2V significantly reduces training and inference costs. 
Extensive experiments across multiple LVU benchmarks also demonstrate its superior accuracy over both general MLLMs and LVU-tailored solutions.
\end{itemize}

\begin{figure*}[!t]        
  \centering  
  \vspace{-0.5cm}        
  \includegraphics[width=1\textwidth]{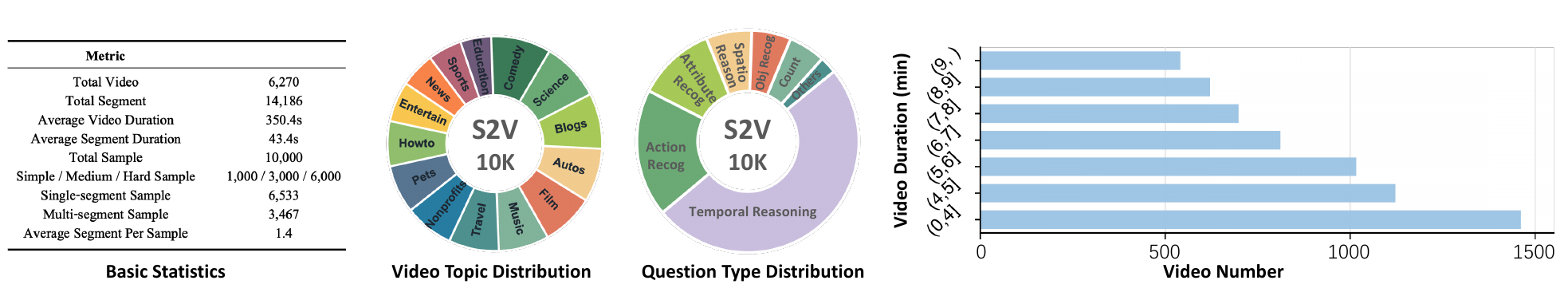}   
  \caption{S2V-10K Statistics and its topic, question type and video duration distribution.}
\label{fig:data_distribution}    
\end{figure*}   

\section{Related Work}
\textbf{General MLLMs for Long Video Understanding.}
Recent MLLMs have achieved remarkable progress in video understanding.
They propose innovative spatio-temporal modeling~\cite{qwen25vl, qwen3vl}, token compression~\cite{minicpm45v}, agentic training~\cite{qwen35, gemini25, gpt5} and mixed reinforcement learning~\cite{internvl35, mimovl7b} to improve video perception and reasoning.
Despite these progress, LVU remains challenging since abundant visual noise in long-form contexts prevent MLLMs from capturing detailed evidence\cite{hu2026spectemp}. 
Our S2V can enhance evidence awareness in MLLMs, mitigating mistakes caused by high visual redundancy in LVU.

\textbf{Retrieval-based Long Video Understanding.}
To obtain useful details in LVU, retrieval-based methods first retrieve query-relevant frames via a tailored retriever model.
And then, the retrieved frames are treated as MLLM input to answer LVU questions. 
These approaches identify relevant contents at once from densely sampled long videos~\cite{abootorabi2025askmodality, liu2025bolt,zhang2025QFrame, yu2025evrag, huang2025frag}, introducing substantial computational and storage burdens due to massive query-frame retrieval. 
Some methods jointly train the retriever and answerer models to bridge their optimization target gap, leading to considerable training overheads~\cite{yao2025gens, yuan2025memvid, kim2025salova, kirolos2024goldfish}.

\textbf{Evidence Reasoning for Long Video Understanding.}
Reasoning-based methods include detailed evidence in MLLM reasoning for long video understanding.
Inspired by Deepseek-R1~\cite{guo2025deepseek}, they leverage a RFT framework to boost reasoning capabilities.
RFT is composed of SFT and RL training, where the former requires large-scale costly COT annotations while the latter involves multiple complicated rewards.
Except for substantial training costs, some methods include the self-reflective and iterative-perception mechanism in the reasoning process, further incurring lengthy outputs and high inference latency. 

Different from these methods, S2V leverages only 10K VQA samples for RL training with a simple VQA accuracy reward and predicts answers using a single MLLM forward pass,
It surpasses existing reasoning-based methods not only in LVU accuracy but also in training and inference efficiency.

\section{Method}
\subsection{S2V Data Construction}

\textbf{Video Collection.}
Since existing evidence-aware LVU datasets are limited by their domain specifically, we regard YT-Temporal-180M~\cite{zellers2021merlot} as S2V video corpus.
This is a large-scale pretraining dataset composed by videos of high quality and spanning diverse domains and topics.
We first retain long-duration videos using the video metadata provided by YT-Temporal-180M. 
And then, we employ a scene detection model to obtain segments according to visual scene transitions.
Such semantic segmentation preserves the visual continuity of each segment, better aligning with realistic video understanding problems.

\textbf{Segment-based and Video-based VQA Generation.}
Given a segment, we instruct the powerful MLLM $M_T$ to generate a VQA triplet in the form of multi-choice question, which consists of question, correct answer, and distracting options.
To reduce hallucinations in $M_T$, we employ another MLLM $M^{\prime}_T$ to re-answer the generated questions and discard VQAs with inconsistent answers.
Except for generating free-form questions, we also allow $M_T$ to generate VQAs following given templates, which are collected from questions of existing video understanding training sets~\cite{tapaswi2016movieqa, lvu2021}. 

Beyond single-segment VQAs, we further generate multi-segment VQAs since cross-segment reasoning is commonly required in LVU~\cite{wang2024videoagent, hu2026spectemp}.
We randomly sample up to $N_M$ segments of each long video as $M_T$ input to generate VQAs the same with single-segment question generation.
Subsequently, we propose a verification step that prompts $M_T$ to recheck if all involved segments are relevant to the generated VQA and finally filter out irrelevant ones.

Up to this point, high-quality segment-based VQA generation has been completed.
However, directly applying these segment-based VQAs to their original whole videos may cause referential ambiguity, as the long video may contain other segments relevant to the same question but associated with different answers.
Therefore, as illustrated in Figure~\ref{fig:framework}, we explicitly incorporate segment timestamp into VQA instruction, guiding models to attend to the corresponding segments.
Such operation effectively transforms segment-based VQAs into exact video-based VQAs conditioned on segments, comprising the final S2V training samples. 

\textbf{Data Filtering.}
Inspired by S1~\cite{muennighoff2025s1}, which highlights the importance of difficult data for LLM post-training, we perform difficulty filtering using a small MLLM $M_I$.
We instruct $M_I$ to respond to each synthesized S2V sample for $N_D$ times. 
Samples with consistently correct answers are regarded as simple, while those with consistently incorrect answers are considered hard. 
The remaining samples, containing both correct and incorrect responses, are categorized as medium difficulty for simple to hard transition.
Since overly simple samples provide limited training benefits, while overly hard samples introduce convergence challenge, we balance simple, medium, and hard samples using ratio determined by experiments.

\begin{table*}[t]\scriptsize
  \begin{center}
    \renewcommand\arraystretch{1.3}
    \setlength\tabcolsep{3pt}
    \begin{tabular}{c|ccccc|cccc|ccccc}
      \Xhline{1px} 
      \multirow{2}{*}{\textbf{Method}} & \multicolumn{5}{c|}{\textbf{LongVideoBench~\cite{wu2024longvideobench}}} & \multicolumn{4}{c|}{\textbf{Video-MME~\cite{fu2025videomme}}} & \multicolumn{5}{c}{\textbf{MLVU~\cite{zhou2024mlvu}}} \\
      \cline{2-15}
      & \textbf{(8,15]} & \textbf{(15,60]} & \textbf{(180,600]} & \textbf{(900,3600]} & \textbf{Overall} & \textbf{Short} & \textbf{Medium} & \textbf{Long} & \textbf{Overall} & \textbf{PlotQA} & \textbf{Needle} & \textbf{Ego} & \textbf{Count} & \textbf{Overall} \\
      \hline 
      \multicolumn{15}{c} {\color{gray}{\textbf{\textit{Closed-Source MLLMs}}}} \\
      \hline
      {\color{gray} GPT-5~\cite{gpt5}} & {\color{gray}76.7} & {\color{gray}72.7} & {\color{gray}62.1} & {\color{gray}56.2} & {\color{gray}63.1} & {\color{gray}84.2} & {\color{gray}72.9} & {\color{gray}67.0} & {\color{gray}74.7} & {\color{gray}80.0} & {\color{gray}78.0} & {\color{gray}65.3} & {\color{gray}38.3} & {\color{gray}72.9} \\
      {\color{gray} Gemini-2.5~\cite{gemini25}} & {\color{gray}76.7} & {\color{gray}83.7} & {\color{gray}70.1} & {\color{gray}61.3} & {\color{gray}69.1} & {\color{gray}86.9} & {\color{gray}79.8} & {\color{gray}72.3} & {\color{gray}79.7} & {\color{gray}82.9} & {\color{gray}79.4} & {\color{gray}65.1} & {\color{gray}60.2} & {\color{gray}78.2} \\
      \hline
      \multicolumn{15}{c} {\textbf{\textit{Open-Source MLLMs}}} \\
      \hline
      Qwen3-VL-4B~\cite{qwen3vl} & 71.4 & 71.5 & 55.6 & 46.8 & 56.2 & 76.3 & 61.3 & 52.6 & 63.4 &72.7 & 70.4 & 60.2 & 37.4 & 66.1 \\
      MiMo-VL-7B~\cite{mimovl7b} & 67.2 & 75.0 & 59.0 & 50.2 & 58.5 & 75.1 & 60.8 & 49.8 & 61.9 & 72.0 & 71.3 & 54.8 & 31.6 & 64.2 \\
      Qwen3-VL-8B~\cite{qwen3vl} & 72.0 & 73.8 & 59.7 & 49.0 & 58.7 & 77.1 & 62.6 & 52.1 & 63.9 & 73.8 & 70.1 & 59.4 & \textbf{51.0} & 69.0 \\
      Qwen2.5-VL-7B~\cite{qwen25vl} & 68.3 & \textbf{76.2} & 60.7 & 49.1 & 58.9 & 74.2 & 61.1 & 52.2 & 62.5 & 68.6 & 77.2 & 56.7 & 32.5 & 65.6 \\
      MiniCPM-V-4.5 (9B)~\cite{minicpm45v} & 70.9 & 72.7 & 60.0 & 48.4 & 58.3 & \underline{77.4} & 64.2 & \textbf{55.1} & \underline{65.6} & 75.1 & 79.2 & 62.8 & 46.6 & 71.0 \\
      Qwen3.5-9B~\cite{qwen35} & 73.0 & 70.3 & \underline{62.9} & 52.3 & 60.8 & 75.7 & 62.9 & \underline{55.0} & 64.5 & \underline{77.9} & 74.9 & 59.4 & 49.0 & 71.8 \\
      InternVL3.5-14B~\cite{internvl35} & 70.4 & 68.6 & 57.8 & 48.8 & 57.1 & 71.0 & 59.2 & 49.1 & 59.8 & 67.0 & 77.7 & 56.0 & 43.7 & 65.6 \\
      \hline
      \multicolumn{15}{c} {\textbf{\textit{Our S2V Models}}} \\
      \hline
      S2V-4B (Ours) & \underline{73.5} & \underline{75.6} & \textbf{63.6} & 53.5 & \underline{62.3} & \textbf{78.2} & 64.1 & 54.4 & \underline{65.6} & \textbf{78.1} & \underline{79.4} & \textbf{68.5} & 44.7 & \underline{71.9} \\
      S2V-7B (Ours) & 72.5 & \underline{75.6} & 61.7 & \textbf{56.0} & \textbf{62.6} & \underline{74.7} & \underline{64.7} & 53.9 & 64.4 & 76.3 & 78.9 & \underline{65.1} & 42.2 & 69.1 \\
      S2V-8B (Ours) & \textbf{74.6} & 75.0 & 60.9 & \underline{55.0} & 62.2 & 77.3 & \textbf{65.7} & 54.2 & \textbf{65.7} & 77.2 & \textbf{79.7} & 62.8 & \underline{50.5} & \textbf{72.0} \\
      \Xhline{1px}
      \end{tabular}
  \caption{Performance comparison results of our S2V \emph{vs.} different general MLLMs on LongVideoBench, Video-MME, and MLVU. Here, best results are in \textbf{bold}, sub-optimal ones are \underline{underlined}.}
  \label{tab:sota_tab_mllm}
  \end{center} 
  \end{table*}

\textbf{Data Statistics and Quality.}
As shown in Figure~\ref{fig:data_distribution}, the resulting S2V-10K dataset consists of 1,000 simple, 3,000 medium, and 6,000 hard samples.
Each sample of S2V-10K contains more than one segment on average, enabling S2V-10K to cover not only needle-in-a-haystack problem but also multi-segment perception and reasoning, all of which are critical LVU challenges.
Figure~\ref{fig:data_distribution} further illustrates the diverse distributions of video duration, topics, and question types of S2V-10K. 
Such diversity allows S2V-10K for comprehensive long video understanding.
 
S2V-10K VQAs are generated from short video segments, allowing a powerful $M_T$ to focus on all fine-grained details.
Therefore, when using $M^{\prime}_T$ to verify the correctness of the generated VQAs, the pass rate can reach 86.7\%. 
With this two-MLLM generation-and-checking pipeline, the final VQAs are of high quality. 
We further manually inspect 500 randomly selected video-level samples.
Benefiting from the introduction of segment timestamp, among which 93.0\% pass the quality check, proving the robustness of our S2V data synthesis pipeline.
We also provide analysis for failure cases in appendix, which are caused by various reasons like challenging target recognition and misleading distractors.

\subsection{S2V Training}
\textbf{Optimization Objective.}
We leverage only RL~\cite{shao2024grpo} for MLLM training based on the question-answering samples provided by S2V-10K.
The training objective is formulated as:
\begin{equation}
  \max_{\theta}\; \mathbb{E}_{(V,Q,I_S,A)\sim \mathcal{D}_{\text{S2V-10K}},\, {A^\prime}\sim \pi_\theta\left({\cdot|V,Q,I_S}\right)}
  \Big[ r(A^\prime,A) \Big],
\end{equation}
where $(V, Q, I_S, A)$ refers to each S2V sample from S2V-10K dataset $\mathcal{D}_{\text{S2V-10K}}$, composed by a whole video $V$, question $Q$, segment location included in the instruction $I_S$ and answer $A$,
$\pi_\theta$ and $A^\prime$ are the policy and predicted answer of the student model $M_S$, reward $r$ is calculated as the question answer accuracy, where correct and incorrect predictions receive rewards of 1 and 0, respectively.

\textbf{Why Does S2V Work?}
It seems that S2V is an easy-to-implement method, which just leverages 10K VQA data for RL training.  
Why can it achieve significant LVU improvement?
Actually, VQA samples generated based on short segments are naturally more likely to involve fine-grained information from localized regions. 
Such details tend to be distracted by other information in long-context videos, thereby overlooked by MLLMs. 
Training on such data can enhance MLLMs in comprehension of fine-grained local evidence, enabling them to better distinguish useful evidence from distracting noise and achieve correct answers.


We have to declare that S2V is not designed for evidence grouding. 
Segment timestamps in S2V are not exact grounding annotations.
They just serve as training-time privileged information to avoid referential ambiguity, which specializes the answer provenance and helps the model to associate local details and correct answer.
This setting is analogous to learning with provenance supervision~\cite{vapnik2015learning}, where privileged information are helpful information to reduce label ambiguity for training, but they are not required information for question answering.
Thus, once the model develops stronger fine-grained video comprehension via training, it can produce correct answers even without any privileged information during inference.

For further provement, we experiment on inference with or without segment timestamp in Experiment section, where the performance difference is small.
We also provide grounding evaluation in appendix. 
Since S2V does not involve exact localization annotation and grounding-targeted training objective, it yields limited improvement.

  \section{Experiments}
  We conduct extensive experiments on multiple typical benchmarks to prove the effectiveness of S2V.
  To be specific, our experiments are intended to answer the following research questions (RQs):

  \begin{itemize}[leftmargin=*, itemsep=0pt, topsep=2pt, parsep=0pt, partopsep=0pt]
    \item \textbf{RQ1} Can S2V improve LVU performance of base MLLMs?
    \item \textbf{RQ2} Can S2V surpass other general MLLMs and LVU-reasoning methods?
    \item \textbf{RQ3} Does S2V enhance fine-grained reasoning for LVU?
    \item \textbf{RQ4} Is S2V data synthesis pipeline effective?
    \item \textbf{RQ5} Does S2V have the potential for boosting LVU performance with larger training scales?
  \end{itemize}

\subsection{Benchmarks and Experimental Settings}
\textbf{Benchmarks and Evaluation Metrics.}
Consistent with existing works~\cite{qwen35,hu2026spectemp}, we experiment on three typical LVU benchmarks, including \textbf{LongVideoBench (LVB)}~\cite{wu2024longvideobench}, \textbf{Video-MME (V-MME)}~\cite{fu2025videomme} and \textbf{MLVU}~\cite{zhou2024mlvu}.
We report accuracy and measure latency of evaluated models for these benchmarks.
More details about them can be obtained in appendix.

  \begin{table}[t]\scriptsize
    \begin{center}
      \renewcommand\arraystretch{1.3}
      \setlength\tabcolsep{3pt}
      \begin{tabular}{c|c|c|c|c|c|c}
        \Xhline{1px} 
        \textbf{Method} & \textbf{\#param} & \textbf{Data}  & \textbf{Frame}  & \textbf{LVB} & \textbf{V-MME} & \textbf{MLVU} \\
        \hline
        \multicolumn{6}{c} {\textbf{\textit{Retrieval-based LVU Methods}}} \\
        \hline
        GoldFish &7B & 176K & -  & - & 28.9 & 37.3 \\
        SALOVA &7B & 2.3M  & 1fps & 44.6  & 53.1 & - \\
        GenS &7B &  150K & 1fps  & 58.7 & - & 66.9 \\
        \hline
        \multicolumn{6}{c} {\textbf{\textit{Reasoning-based LVU Methods}}} \\
        \hline
        Video-R1 &7B & 425K & 64 & 58.5 & 61.4 & 62.4 \\
        VideoRFT &7B & 412K & 64 & 58.6 & 61.5 & 63.9 \\
        VideoChat-R1.5  &7B & 80K & 64  & 60.6 & 63.4 & 52.3 \\
        VIDEOP2R  &7B & 162K& 32 & - & 60.0 & - \\
        SpecTemp &7B & 80K & 48 &61.4 & 64.1 & 50.9  \\
        FrameThinker &7B & 30K & 22 & 52.9 & - & 59.1 \\
        Open-o3-Video &7B & 66K & 64 & - & 63.6 & - \\
        VideoMind  &7B & 481K & 399 & 56.3 & 58.2 & 64.4 \\
        \hline
        \multicolumn{6}{c} {\textbf{\textit{Our S2V Models}}} \\
        \hline
        S2V (Ours) & 4B & 10K & 64 & \underline{62.3} & \underline{65.6} & \underline{71.9} \\
        S2V (Ours) & 7B & 10K & 64 & \textbf{62.6} & 64.4 & 69.1 \\
        S2V (Ours) & 8B & 10K & 64 & 62.2 & \textbf{65.7} & \textbf{72.0} \\
        \Xhline{1px}
        \end{tabular}
    \caption{Performance comparison results of our S2V \emph{vs.} different LVU-tailored methods on LongVideoBench, Video-MME, and MLVU. Here, \textbf{Data} refers to training data scale and \textbf{Frame} denotes the input frame number.}
    \label{tab:sota_tab_lvu}
    \end{center} 
    \end{table}

\textbf{Implementation Details.}
For data synthesis, we employ Qwen3-VL-235B as $M_T$ for VQA generation.
The maximum segment number $N_M$ for multi-segment VQA generation is 3.
We leverage GLM-4.5V~\cite{vteam2025glm45v} as the VQA recheck model $M_T^\prime$ and Qwen3-VL-8B~\cite{qwen3vl} as the difficulty filtering model $M_I$ with 4 as the response repetition number $N_D$.
We uniformly sample 64 frames to represent the whole video, while frame groups corresponding to the detected segments are used to represent segments. 
Frame indices of these detected segments within the 64 sampled frames are treated as segment timestamp.
Notably, to avoid data leakage, we compare our S2V-10K videos with all evaluation benchmarks and confirm no overlap.

We leverage Group Relative Policy Optimization (GRPO)~\cite{shao2024grpo} implemented by ms-swift~\cite{zhao2025swift} to train S2V-4B, S2V-7B and S2V-8B, whose parameters are initialized from Qwen3-VL-4B~\cite{qwen3vl}, MiMo-VL-7B~\cite{mimovl7b} and Qwen3-VL-8B~\cite{qwen3vl}.
We conduct training on 8 80G A100 GPUs for 1 epoch with 8 GRPO rollouts.
Except for the data-scaling experiment, all experimental results are obtained based on S2V-10K.
More training details and all prompts are provided in appendix.
  
\textbf{Comparison Baselines.}
S2V is evaluated against plentiful methods for a comprehensive comparison, which includes:
1) state-of-the-art (SOTA) general MLLMs with comparable parameter scales to S2V, covering Qwen3-VL-4B~\cite{qwen3vl}, MiMo-VL-7B~\cite{mimovl7b}, Qwen3-VL-8B~\cite{qwen3vl}, Qwen2.5-VL-7B~\cite{qwen25vl}, MiniCPM-V-4.5 (9B)~\cite{minicpm45v}, Qwen3.5-9B~\cite{qwen35} and InternVL3.5-14B~\cite{internvl35};
and 2) LVU-tailored methods covering both retrieval-based and reasoning-based methods, where GoldFish~\cite{kirolos2024goldfish}, SALOVA~\cite{kim2025salova} and GenS~\cite{yao2025gens} are retrieval-based methods, Video-R1~\cite{feng2025videor1}, VideoRFT~\cite{wang2025videorft}, VideoChat-R1.5~\cite{yan2025videochatr15}, VIDEOP2R~\cite{jiang2026videop2r}, SpecTemp~\cite{hu2026spectemp}, FrameThinker~\cite{he2025framethinker}, VideoMind~\cite{liu2026videomind} and Open-o3-Video~\cite{meng2025openo3} are reasoning-based methods.

\begin{figure}[!t]        
  \centering        
  \includegraphics[width=1\columnwidth]{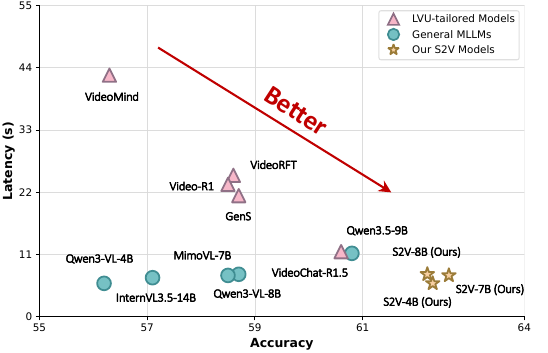}     
  \caption{Performance and latency comparison results of our S2V \emph{vs.} both general MLLMs and LVU-reasoning methods on LongVideoBench benchmark.}
\label{fig:sota_tab_latency}   
\end{figure} 

\begin{figure*}[!t]         
  \centering        
  \includegraphics[width=1\textwidth]{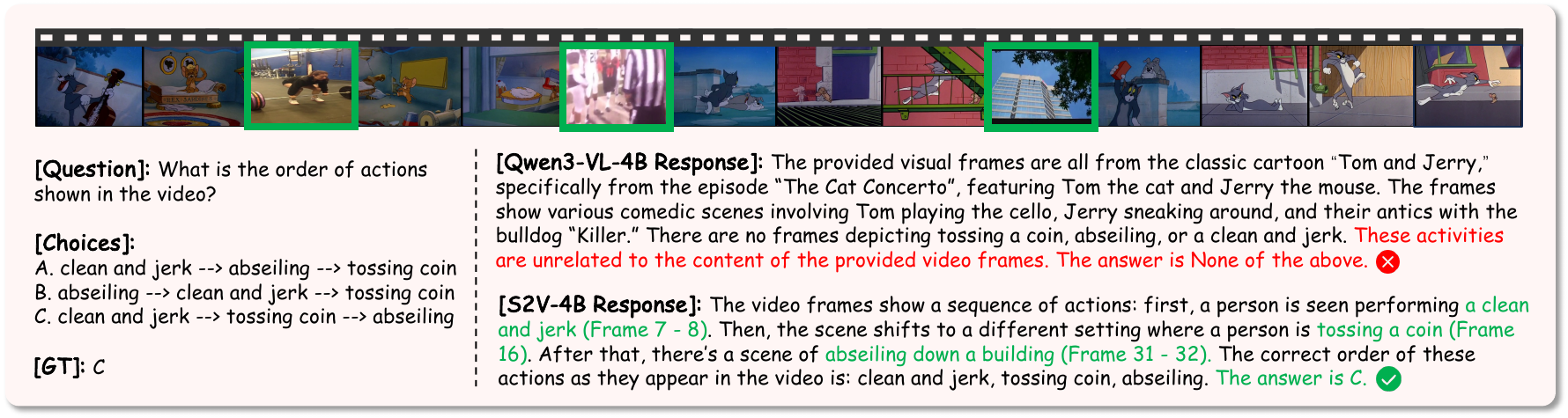}     
  \caption{Example LVU results of MLVU predicted by Qwen3-VL-4B \emph{vs.} S2V-4B. Here, query-relevant frames are marked with green boxes.}
  \label{fig:case_study} 
  \end{figure*}

\begin{figure}[!t]        
  \centering        
  \includegraphics[width=1\columnwidth]{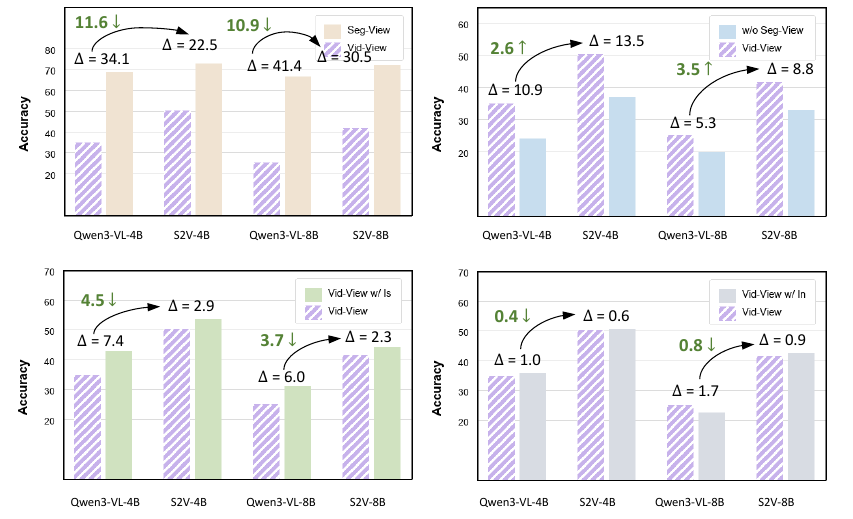}     
  \caption{Performance comparison results of our S2V \emph{vs.} initial MLLMs. Here, \textbf{Seg-View}, \textbf{Vid-View} and \textbf{w/o Seg-View} refer to leveraging the corresponding segment, whole video and the video with the corresponding segment removed to answer the question respectively, \textbf{Vid-View w/ Is} and \textbf{Vid-View w/ In} leverage the whole video with segment position or null position in instruction to answer question respectively, and \textbf{$\Delta$} is the VQA accuracy gap.} 
  \label{fig:ablation_s2vgap}     
\vspace{-0.5cm}   
\end{figure}

\subsection{SOTA Comparison (RQ1 \& RQ2)}
\textbf{Comparison with General MLLMs.}
As Table~\ref{tab:sota_tab_mllm} shows, compared to base MLLMs, S2V can consistently improve the LVU performance across all LVU benchmarks, further outperforming SOTA MLLMs.
Notably, compared with the short-video subsets, S2V achieves more pronounced gains on the medium- and long-video subsets of LongVideoBench, demonstrating its effectiveness for LVU.
Moreover, S2V achieves strong performance on the needle-in-a-haystack (Needle) and counting (Count) subsets of MLVU, explicitly demonstrating its enhanced localized reasoning over both single and multiple segments in LVU.

\textbf{Comparison with Methods Tailored for LVU.}
Table~\ref{tab:sota_tab_lvu} verifies that S2V surpass both retrieval-based and reasoning-based methods in LVU accuracy.
With only 10K training samples, S2V achieves superior accuracy over methods that rely on substantially larger datasets.
Compared to Video-R1, it reduces training data from 425K to 10K samples (98\% reduction) and GPU hours from approximately 440 to 160 (64\% reduction), achieving remarkable training efficiency.
Notably, S2V-7B builds on MiMo-VL-7B, which in Table~\ref{tab:sota_tab_mllm} is a relatively weaker 7B backbone than Qwen2.5-VL-7B, and the latter is the backbone of most LVU-tailored methods.
This indicates that the improvement of S2V is not due to backbone advantage.

\textbf{Latency Comparison.}
Figure~\ref{fig:sota_tab_latency} highlights the inference efficiency of S2V, where S2V achieves better accuracy-latency tradeoff than both general MLLM and reasoning-based baselines.
Benefiting from the effectiveness of S2V, even the lightweight S2V-4B achieves significant training gains, achieving high LVU accuracy with low inference cost.
And different from some reasoning-based methods that require iterative perception, S2V enables models to obtain accurate answers through a single forward pass, significantly reducing inference latency.

\begin{figure}[!t]        
  \centering        
  \includegraphics[width=1\columnwidth]{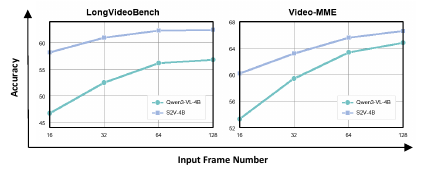}     
  \caption{Ablation study of taking different numbers of frames as input on LongVideoBench and Video-MME.} 
  \label{fig:ablation_frame_number}     
  \end{figure}

\subsection{Fine-grained Reasoning Enhancement (RQ3)}
\label{sec:localized_reasoning}     
To verify S2V can enhance reasoning via percepting and comprehending fine-grained evidence, we visualize S2V reasoning output in Figure~\ref{fig:case_study}. 
The given question asks about the temporal order of few realistic actions sparsely interspersed throughout the long cartoon video.
The base MLLM prematurely focuses on the primary ``Tom \& Jerry'' storyline, completely overlooking the query-relevant actions and incorrectly predicting ``None''.
In contrast, our S2V model exactly captures all involved actions, clearly indicating their locations and achieving the correct answer ``C''.
More cases in appendix can further support this observation.

Additionally, we collect a S2V-Test set composed of 800 samples to verify S2V indeed enhances evidence awareness.
As observed in upper half of Figure~\ref{fig:ablation_s2vgap}, QA accuracy evaluated on segments (Seg-View) distinctly outperforms those on whole videos (Vid-View), which is intuitive since the corresponding segment inherently includes prominent query-relevant information while the whole video contains more misleading noise.
We calculate performance gap between Seg-View and Vid-View, which is significantly reduced after S2V training.
We conjecture that S2V contributes to perceiving and correctly utilizing evidence in segments, leading to Vid-View results more aligned with Seg-View results.
We also report results predicted on videos with the corresponding segments removed (w/o Seg-View), where all models show accuracy degradation.
Notably, S2V models exhibit larger performance drops than their base models, indicating that S2V relies more strongly on localized evidence when answering questions.

As shown in the lower half of Figure~\ref{fig:ablation_s2vgap}, for the base model, segment timestamps in instruction provides a beneficial guide. 
However, after S2V training, the performance gap between inference with (Vid-View w/ Is) and without segment timestamps (Vid-View) becomes much smaller. 
S2V improves the perception and comprehension of localized details, reducing its reliance on explicit timestamp cues, for which the train-inference mismatch has only a limited impact on S2V.
We also preserves the segment guide in instruction but provides only a null timestamp (Vid-View w/ In). 
This setting achieves similar result to that without any segment timestamp (Vid-View), indicating that instruction of timestamp-guide style cannot lead to S2V improvement.

Figure~\ref{fig:ablation_frame_number} shows the experimental results of using different input frame numbers for LVU inference.
Compared to the base MLLM, the improvement of S2V consistently exists across all input frame numbers even though the video frame number of S2V-10K is fixed as 64, confirming the robustness of S2V.
More importantly, compared with general MLLMs, whose performance varies significantly with the number of input frames, S2V exhibits much smaller performance fluctuations.
We attribute this to a shift from ``looking more'' to ``looking better'': general MLLMs rely on dense visual inputs to confirm key evidence, whereas S2V can accurately comprehend sparse query-relevant clues, enabling effective LVU with fewer input frames, significantly mitigating visual redundancy.

\begin{table}[t]\scriptsize
  \begin{center}
    \renewcommand\arraystretch{1.3}
    \setlength\tabcolsep{3pt}
      \begin{tabular}{
        >{\centering\arraybackslash}m{2.15cm}|
        >{\centering\arraybackslash}m{0.9cm}|
        >{\centering\arraybackslash}m{0.9cm}|
        >{\centering\arraybackslash}m{0.9cm}|
        >{\centering\arraybackslash}m{0.65cm}|
        >{\centering\arraybackslash}m{0.65cm}
      }
      \Xhline{1px} 
      \textbf{Method} & $\mathbf{VQA}_{\mathbf{V}}$ & $\mathbf{VQA}_{\mathbf{S}}$ & $\boldsymbol{I}_{\boldsymbol{S}}$ & \textbf{LVB} & \textbf{MLVU} \\
      \hline
      Qwen3-VL-4B & \xmark & \xmark & \xmark  & 56.2 & 66.1 \\
      \hline
      V2V-4B  & \cmark & \xmark & \xmark  & 57.7 & 67.0 \\
      S2V-4B w/o loc  & \cmark & \cmark & \xmark  & \underline{60.6}  & \underline{69.3} \\
      S2V-4B & \cmark & \cmark & \cmark & \textbf{62.3}& \textbf{71.9} \\
      \Xhline{1px}
      \end{tabular}
  \caption{Ablation study of our S2V synthesis pipeline on LongVideoBench and MLVU. Here, $\mathbf{VQA}_{\mathbf{V}}$ and $\mathbf{VQA}_{\mathbf{S}}$ refer to VQA annotation based on the whole video and segments, respectively and $\boldsymbol{I}_{\boldsymbol{S}}$ is the segment timestamp included in the VQA instruction.} 
  \label{tab:ablation_data_synthesis}
  \end{center} 
  \end{table} 
   
\begin{figure}[!t]        
  \centering        
  \includegraphics[width=1\columnwidth]{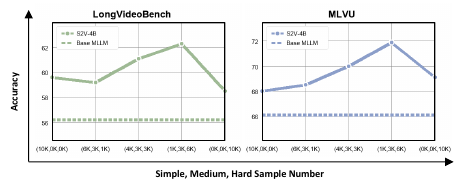}     
  \caption{Ablation study of simple, medium and hard sample numbers in S2V-10K on LongVideoBench and MLVU.}
  \label{fig:ablation_data_sample}    
  \vspace{-0.3cm}  
  \end{figure}    

\subsection{Ablation Study (RQ4 \& RQ5)}

\textbf{Data Synthesis Pipeline.}
To demonstrate the effectiveness and rationality of our S2V data synthesis pipeline, we implement two alternative synthesis strategies, as shown in Table~\ref{tab:ablation_data_synthesis}.
One generates VQAs directly from the whole video (V2V), while the other adapts segment-based VQAs on the whole video without any segment timestamp (S2V w/o loc).
V2V can be regarded as the conventional supervision that make annotations from the whole video LVU training. 
Compared with the initial model, training on V2V data yields only limited improvement, suggesting that general video VQAs are insufficient for current MLLMs, which have been extensively trained on large-scale whole-video oriented LVU datasets.
The performance of S2V w/o loc lies between V2V and the complete S2V, from which we can conclude that:
1) even without specifying corresponding segments, applying segment-based VQA on the whole video still encourages model to perceive query-relevant fine-grained evidence;
and 2) generated questions contain ambiguous segment references, for which removing segment timestamp achieves sub-optimal training performance.

\textbf{Simple, Medium and Hard Sample Number.}
We also investigate the influence of simple, medium, and hard samples in S2V-10K. 
As shown in Figure~\ref{fig:ablation_data_sample}, LVU performance improves as the number of hard samples increases, until all simple and medium samples are removed.
By analyzing the training logs, we find that only simple samples lead to rapid convergence but yields sub-optimal LVU improvement, while only hard samples cause optimization instability and convergence challenges. 
Therefore, we adopt a balanced simple-medium-hard ratio of 1:3:6 in S2V-10K. 
Moreover, across all simple-medium-hard ratios, S2V consistently improves over the base MLLM, suggesting that the advantage of S2V mainly stems from the training paradigm of segment-to-video supervision, rather than merely difficult sample filtering.

    \begin{figure}[!t]        
      \centering  
      \includegraphics[width=1\columnwidth]{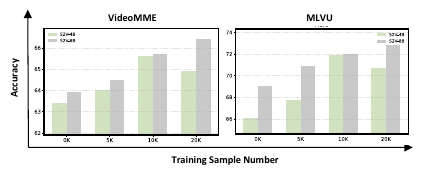}   
      \caption{Ablation study of S2V trained with different data scales and model scales on Video-MME and MLVU.}
    \label{fig:ablation_training_scale}    
    \end{figure}

\textbf{Training Scale.} 
To explore the scalability of S2V, we train 4B and 8B models with training samples of different scales. 
As shown in Figure~\ref{fig:ablation_training_scale}, S2V-8B consistently outperforms S2V-4B under all data scales, verifying the potential of applying S2V to larger-scale models. 
Moreover, the LVU performance of S2V-8B further improves with more training data, while that of S2V-4B slightly drops. 
We conjecture that larger-scale models have greater capacity and more room for improvement, making them better benefit from additional S2V data and handle challenging LVU problems.

\section{Conclusion}
In this paper, to address the substantial training and inference overheads in existing reasoning-based methods, we proposed a novel, effective and efficient \emph{\textbf{S}egment-\textbf{t}o-\textbf{V}ideo Supervision} method.
It first generated VQAs based on short segments and then transferred this segment-based VQAs back to the whole video. 
Training on such data enforced models to associate fine-grained evidence with QA while avoiding distracting noise in the whole video.
The efficient S2V involves just reinforcement learning (RL) with a simple accuracy reward based on only 10K VQA samples and predicts answer using a single forward pass with limited output tokens.
Experimental results across multiple benchmarks demonstrated that S2V can consistently improve the LVU performance, outperforming general MLLMs and reasoning-based methods in both LVU accuracy and training and inference efficiency.

\bibliography{aaai2027}

\appendix
\newpage
\section{Appendix}
In this supplementary material, we provide the following: 1) details of evaluation benchmarks; 2) S2V training details; 3) experiment for video grounding; 4) ablation study of using RL \emph{vs.} RFT for S2V training; and 5) comprehensive case study, including S2V-10K sample, failed data construction examples and S2V reasoning outputs.

\subsection{Evaluation Benchmark Details} 
\label{sec:benchmark_details} 
\textbf{LongVideoBench}~\cite{wu2024longvideobench} contains 6,678 multiple-choice questions on 3,763 videos.
These videos are diverse in their themes, including movies, news, life and knowledge, covering 4 progressive duration groups: 8-15 seconds, 15-60 seconds, 3-10 minutes, and 15-60 minutes.
Questions are divided into 17 fine-grained categories, with human-annotated choices, covering a wide range of video understanding tasks.
In this paper, we take its validation set for evaluation.

\textbf{Video-MME}~\cite{fu2025videomme} annotates 2,700 high-quality multiple-choice questions on 900 videos covering six key domains: knowledge, film \& television, sports competition, life record, and multilingual.
Each domain involves videos with varying duration lengths, including short (< 2 minutes), medium (4-15 minutes), and long videos (30-60 minutes), yielding a median certificate length of 26s, 164.7s, and 890.7s for short, medium and long videos, respectively.

\textbf{MLVU}~\cite{zhou2024mlvu} consists of 3,102 questions on 1,730 videos across 9 categories, specifically designed for long video understanding. 
These videos range from 3 minutes to 2 hours and the average video length is about 15 minutes.
MLVU includes both multiple-choice (MCQ) and open-ended generation questions and in this paper, we take its MCQ set for S2V evaluation. 
Except for the overall performance, we particularly report performance of \textbf{Plot-QA}, \textbf{Needle}, \textbf{Ego} and \textbf{Count} subsets of MLVU, which are more related to temporal perception in LVU, explicitly demonstrating the grounding enhancement of S2V.

\subsection{S2V Training Details}
\label{sec:training_details}
We employ LoRA for efficient GRPO training. 
We set the LoRA Rank to 32, GRPO rollout group to 8, rollout temperature to 1.0, train S2V on 8 80G A100 GPUs for 1 epoch using AdamW Optimizer and set the batch size to 64, learning rate to 1 {×} ${10}^{-5}$.
The maximum image pixel for training and inference are 50,176 and 174,080, respectively.
All Qwen-3-VL models used in this paper are ``Instruct'' versions.

\begin{figure*}[!t]         
  \centering        
  \includegraphics[width=1\textwidth]{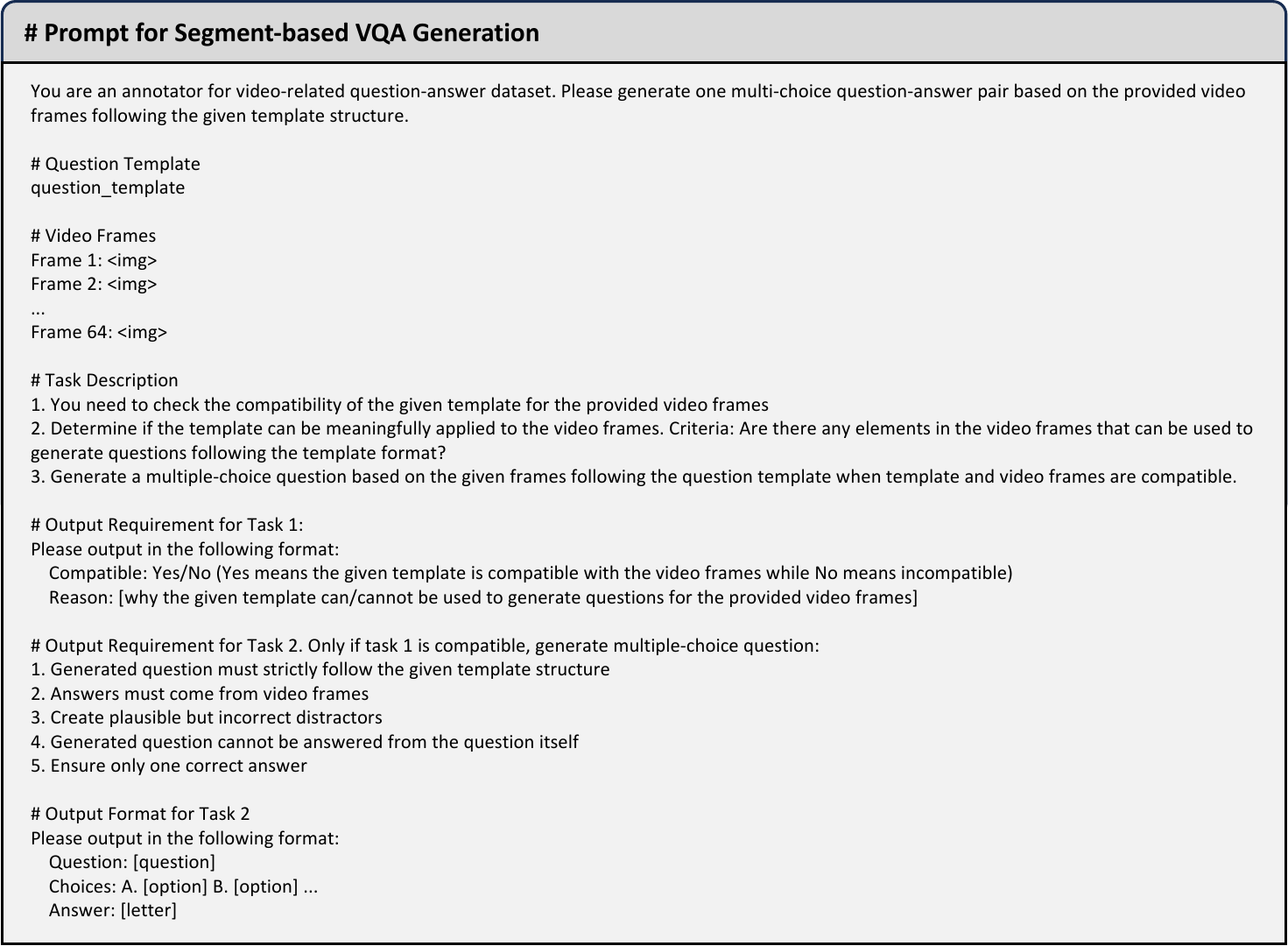}     
  \caption{The segment-based VQA generation prompt in the segment-based VQA generation step of S2V-10K. This prompt guides the model to generate a multi-choice question answer triplet including question, answer and distracting options based on the provided segment frames and question template.}
  \label{fig:vqa_generation_prompt} 
  \end{figure*}

\begin{figure*}[!t]         
    \centering        
    \includegraphics[width=1\textwidth]{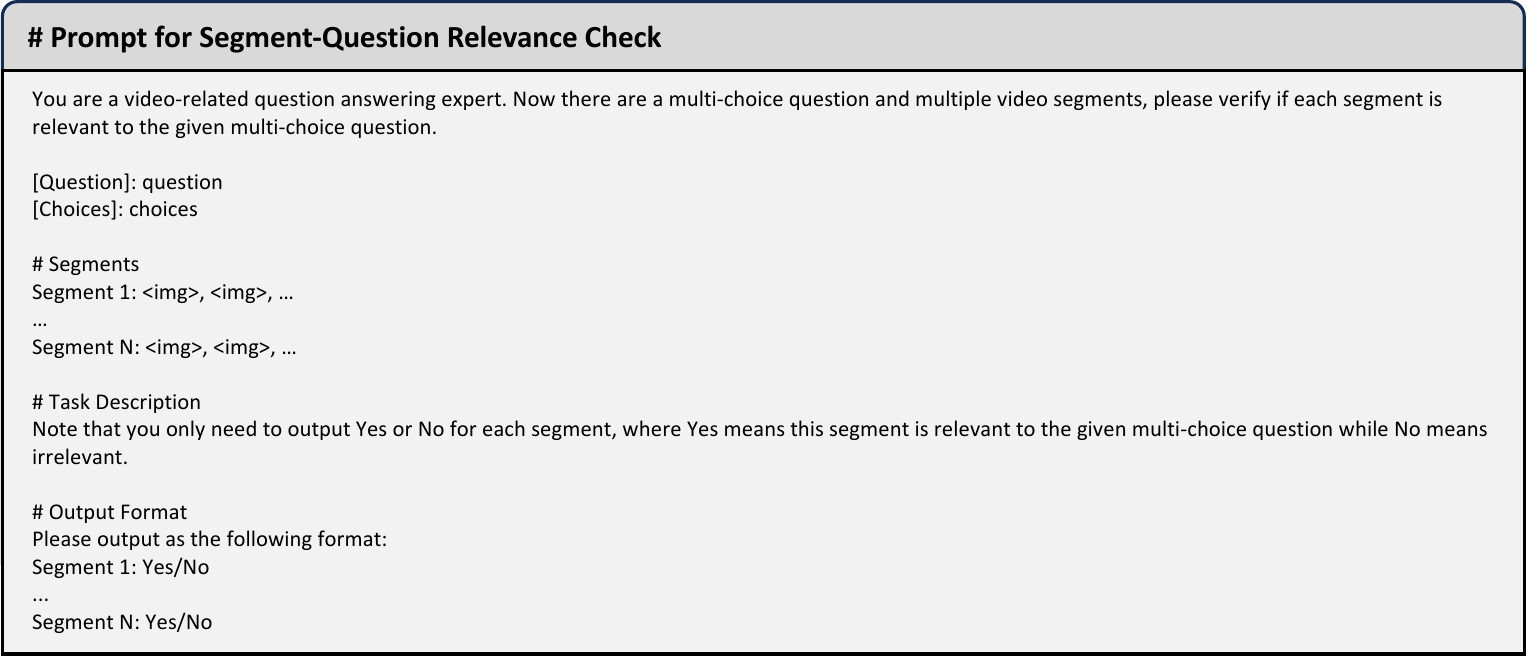}     
    \caption{The segment-question relevance check prompt in the multi-segment VQA generation step of S2V-10K. This prompt guides the model to check if each segment involved in the multi-segment sample is truly relevant to the generated VQA.}
    \label{fig:segment_relevance_prompt} 
    \end{figure*}

\label{sec:prompts}
To enhance reproducibility and transparency, we provide all prompts involved in the S2V data synthesis pipeline.
Figure~\ref{fig:vqa_generation_prompt} is the segment-based VQA generation prompt in the segment-based VQA generation step of S2V-10K.
Figure~\ref{fig:segment_relevance_prompt} shows the segment-question relevance check prompt in the multi-segment VQA generation step of S2V-10K.

\subsection{Video Grounding Evaluation}
S2V improves fine-grained reasoning for long video understanding, which raises the question of whether it can achieve more precise evidence grounding.
We conduct this experiment using TVGBench~\cite{wang2025timer1}. 
This is a lightweight yet comprehensive evaluation benchmark specifically designed for temporal video grounding.
It contains 800 instances with balanced video durations, covering 11 semantic categories under three major types: human, object, and environment.

As shown in Table~\ref{tab:sota_tab_tvg}, experimental results demonstrate that S2V consistently achieves enhanced grounding capability.
However, the performance gain is not particularly significant.
We think this is because S2V annotations do not involve exact evidence positions and there are no grounding-related optimization objectives during S2V training.

\subsection{RL \emph{vs.} SFT}

S2V involves only multiple-choice question (MCQ) annotations, where the sole supervision signal is the correct option label, such as A, B, or C. 
We argue that such single-letter signals contain little semantic information and unsuitable for SFT.
Therefore, we adopt only RL for S2V training. 
The results in Table~\ref{tab:ablation_rl} further support this choice: 
when trained on the same S2V-10K, LVU performance of using SFT is substantially lower than that using RL, where SFT even degrades LVU performance compared to base MLLM.

\begin{table}[t]\scriptsize
  \begin{center}
    \renewcommand\arraystretch{1.3}
    \setlength\tabcolsep{3pt}
    \begin{tabular}{c|c|c|c|c}
      \Xhline{1px} 
      \textbf{Method} & \textbf{\#param} & \textbf{Recall@0.3} & \textbf{Recall@0.5} & \textbf{mIoU} \\
      \hline
      TimeChat & 7B & 22.4 & 5.3  & - \\
      Qwen2.5-VL & 7B & 25.9 & 7.8  & 17.8 \\
      Qwen3-VL & 4B &26.0 & 6.9 &  17.7 \\
      MiMo-VL & 7B & \underline{27.9} & \underline{9.3}  & \underline{19.1} \\
      Qwen3-VL & 8B &25.4 & 9.0 &  17.0 \\
      \hline
      S2V & 4B & 27.0 & 7.1 &  18.4 \\
      S2V & 7B & \textbf{29.0} & \textbf{9.9}& \textbf{19.9} \\
      S2V & 8B   & 26.8 & \underline{9.3} & \underline{19.1} \\
      \Xhline{1px}
      \end{tabular}
  \caption{Performance comparison results of our S2V \emph{vs.} different general MLLMs and TVG-tailored methods on TVGBench.}
  \label{tab:sota_tab_tvg}
  \end{center} 
  \vspace{-0.5cm}
  \end{table}

  \begin{table}[t]\scriptsize
    \begin{center}
      \renewcommand\arraystretch{1.3}
      \setlength\tabcolsep{3pt}
      \begin{tabular}{c|c|c|c|c|c}
        \Xhline{1px} 
        \textbf{Method} & \textbf{SFT} & \textbf{RL} & \textbf{LongVB} & \textbf{Video-MME}& \textbf{MLVU} \\
        \hline
        Qwen3-VL-4B & \xmark & \xmark & 56.2 & 63.4  & 66.1 \\
        S2V-4B  & \cmark & \xmark & 55.6 &61.1 & 67.0 \\
        S2V-4B & \xmark & \cmark  & 62.3 & 65.6 & 71.9 \\
        \hline
        MiMo-VL-7B & \xmark & \xmark & 58.5 & 61.9  & 64.2 \\
        S2V-7B  & \cmark & \xmark & 52.3 & 56.0 & 61.6 \\
        S2V-7B & \xmark & \cmark  & 62.6 & 64.4 & 69.1 \\
        \hline
        Qwen3-VL-8B & \xmark & \xmark & 58.7 & 63.9  & 69.0 \\
        S2V-8B  & \cmark & \xmark & 56.3 & 61.5 & 68.0 \\
        S2V-8B & \xmark & \cmark  & 62.2 & 65.7 & 72.0 \\
        \Xhline{1px}
        \end{tabular}
    \caption{Ablation study of S2V using different training strategies on LongVideoBench (LongVB), Video-MME and MLVU.}
    \label{tab:ablation_rl}
    \end{center} 
    \vspace{-0.5cm}
    \end{table}

\subsection{Case Study}
\begin{figure*}[t]
  \centering
  \includegraphics[width=1\textwidth]{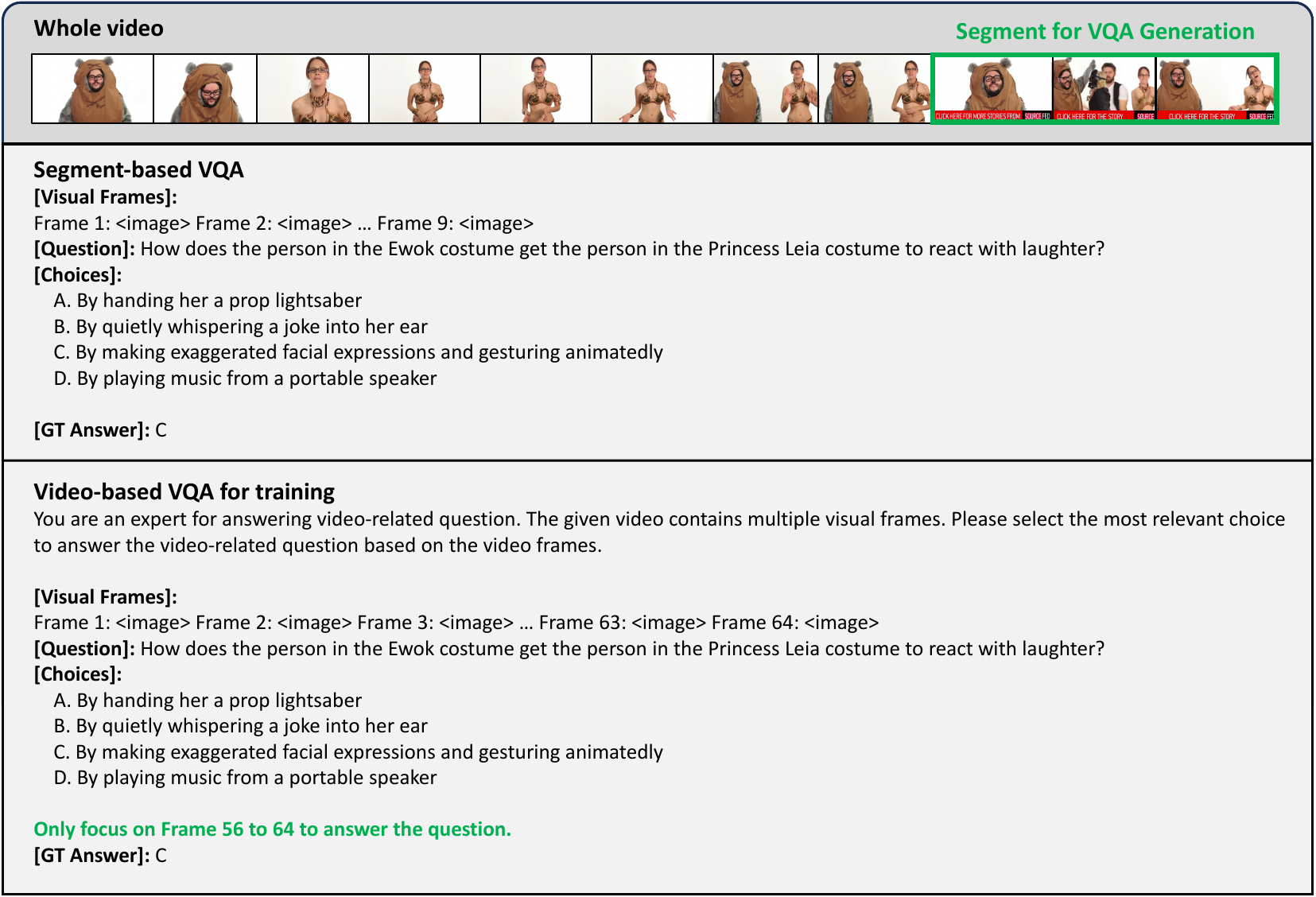}
  \caption{A segment-based VQA example and the corresponding converted video-based VQA sample in S2V-10K.}
  \label{fig:s2v10k_example}
\end{figure*}

\textbf{S2V-10K Sample.}
To better understand the S2V data, we visualize an S2V sample in Figure~\ref{fig:s2v10k_example}, which shows a segment-based VQA example and its converted video-based VQA for S2V-10K.
By incorporating the corresponding segment location, we can efficiently obtain a video-based VQA from a segment-based VQA.

\begin{figure*}[t]
  \centering
  \includegraphics[width=1\textwidth]{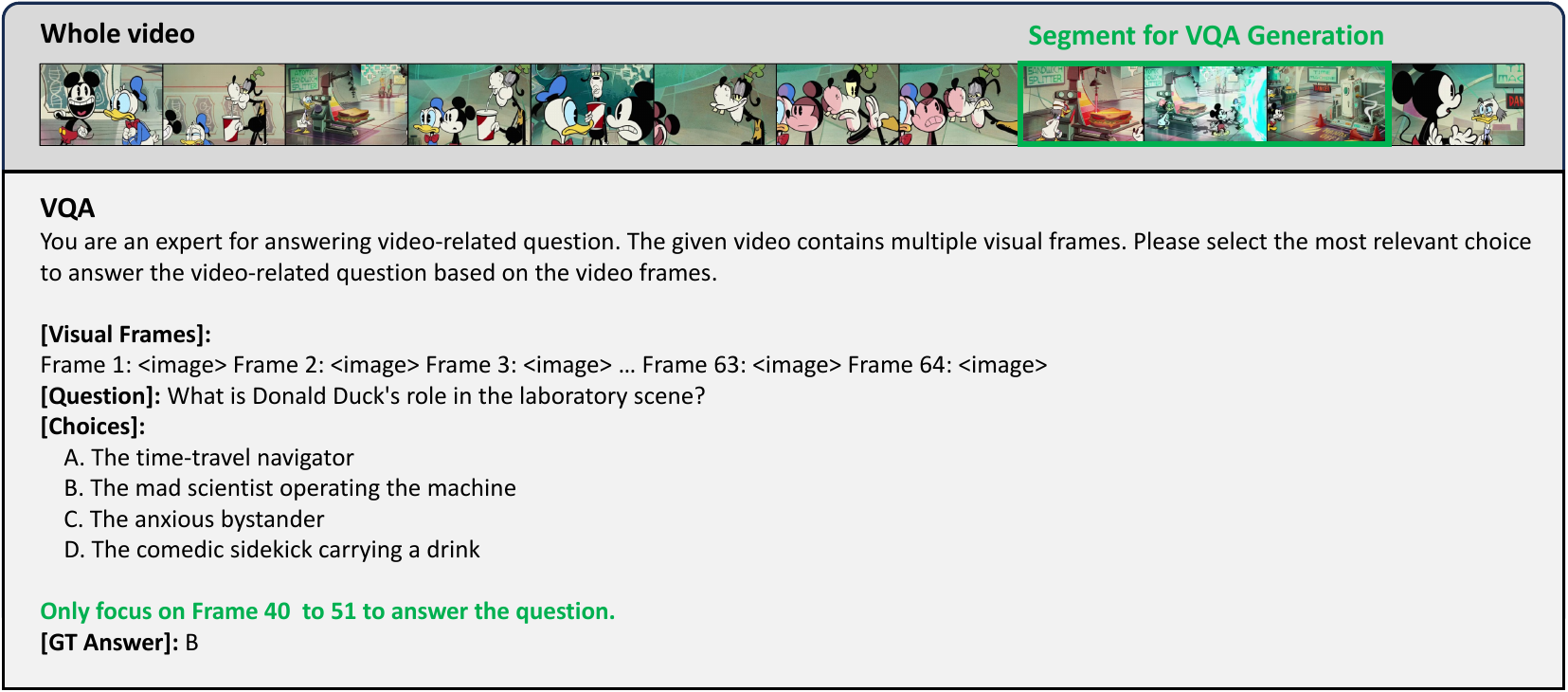}
  \caption{A failed data construction case caused by object recognition error.}
  \label{fig:fail_donad}
\end{figure*}

\begin{figure*}[t]
  \centering
  \includegraphics[width=1\textwidth]{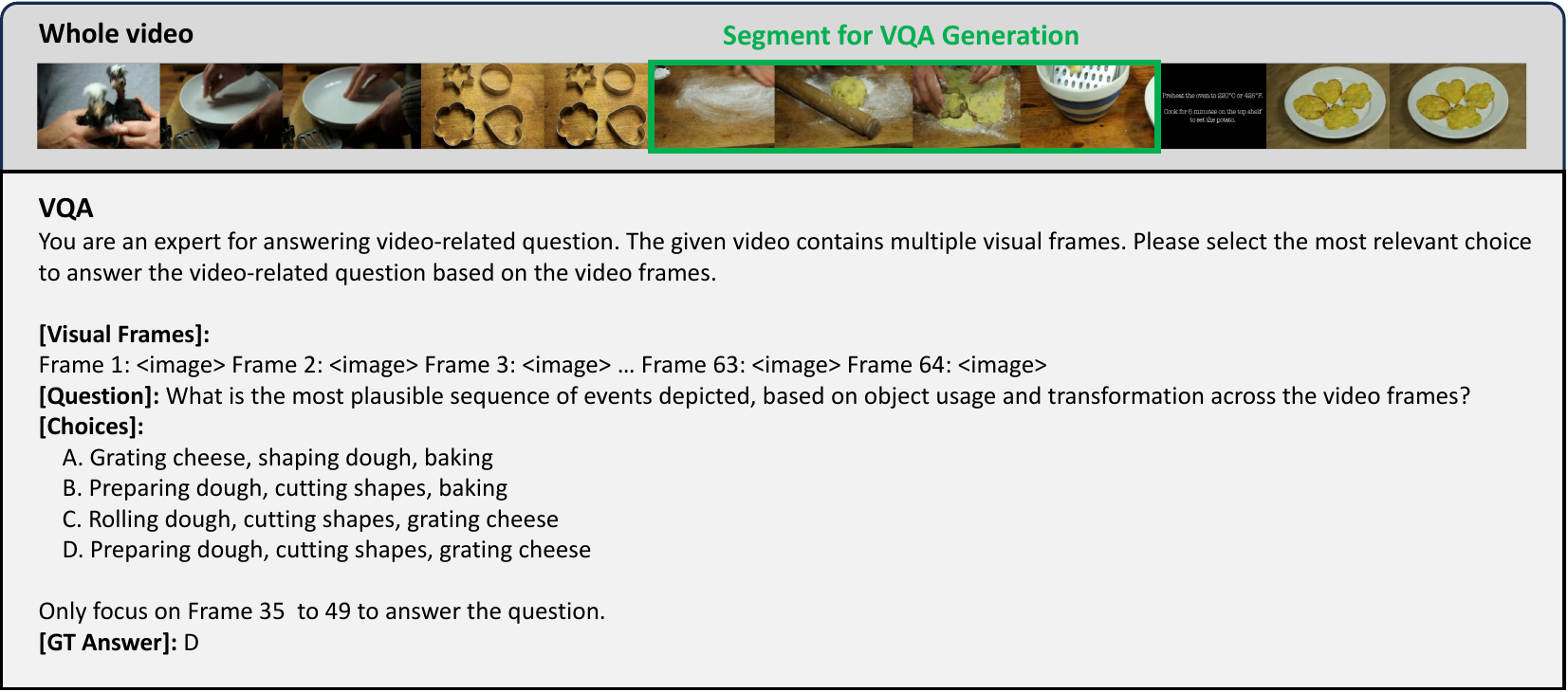}
  \caption{A failed data construction case caused by confusing distractors.}
  \label{fig:fail_baking}
\end{figure*}

\begin{figure*}[t]
  \centering
  \includegraphics[width=1\textwidth]{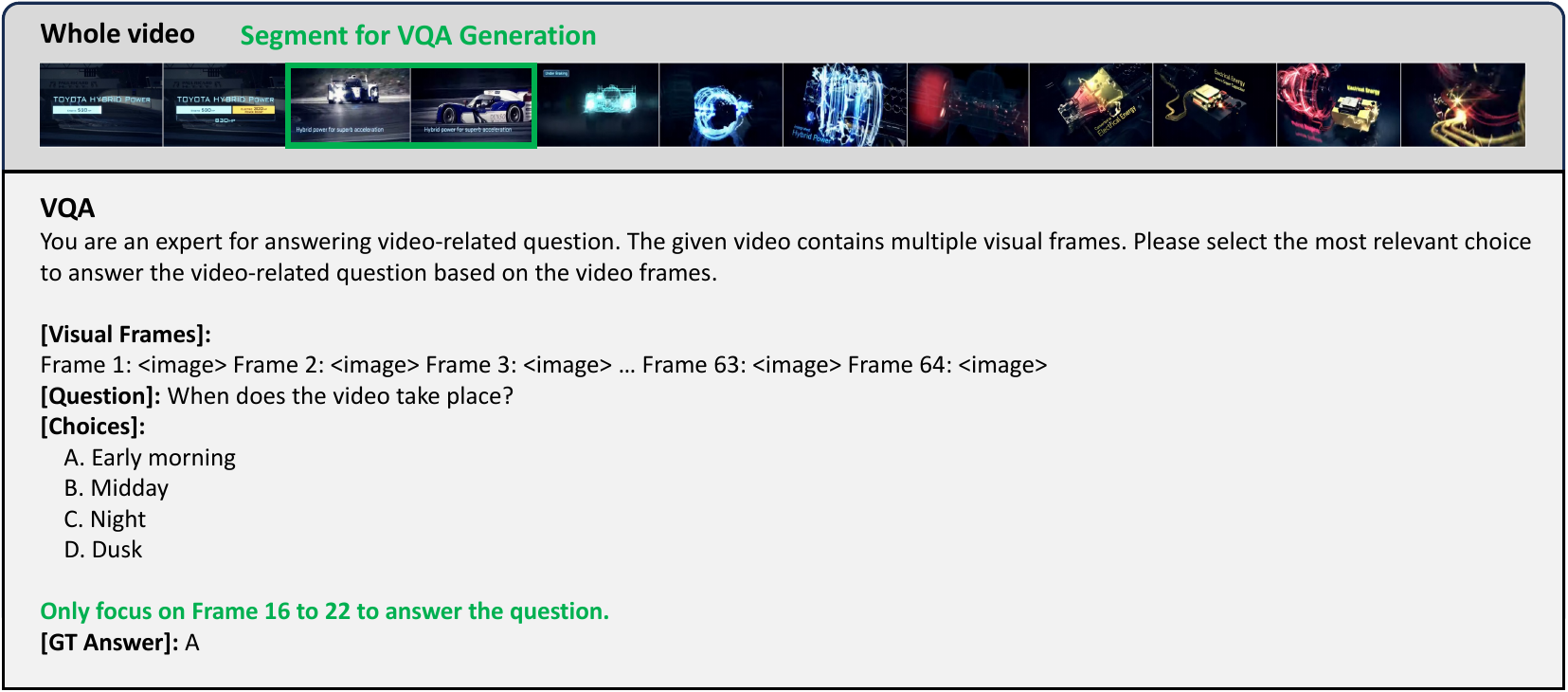}
  \caption{A failed data construction case caused by ambiguity in segment.}
  \label{fig:fail_car}
\end{figure*}

\textbf{Failed Data Construction Examples.}
Although the automated construction pipeline of S2V is robust, achieving a human pass rate of up to 93\%, there still exist some failed construction cases caused by various reasons.
Figure~\ref{fig:fail_donad} shows an object recognition error, where the model mistakenly identifies Professor Ludwig Von Drake as Donald Duck, resulting in an incorrect correspondence between the entity in the question and the answer. 
Figure~\ref{fig:fail_baking} illustrates confusing distractor options: options C and D are identical except for the first dough-related step, while rolling out dough can be regarded as a subset of preparing dough, making the two options difficult to distinguish. 
Figure~\ref{fig:fail_car} further shows ambiguity within a segment, where both early-morning and nighttime scenes appear in the same segment, making both options A and C valid.

\begin{figure*}[t]
  \centering
  \includegraphics[width=1\textwidth]{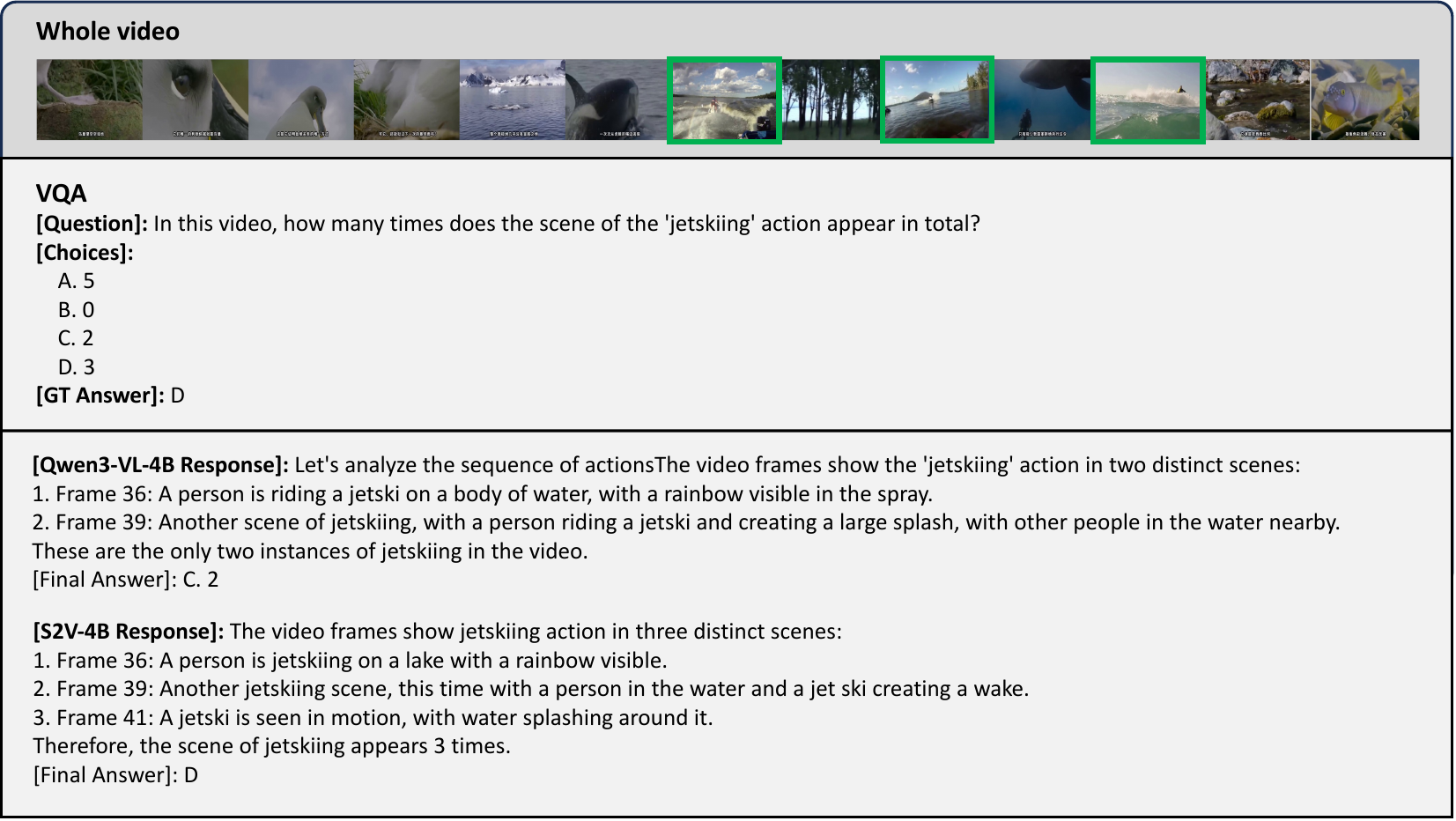}
  \caption{An S2V prediction output includes localized details which tend to be overlooked.}
  \label{fig:example_jetsking}
\end{figure*}

\textbf{S2V Reasoning Outputs.}
We visualize several prediction results of S2V, showing that its reasoning enhances fine-grained evidence understanding from two aspects.
First, S2V can include fine-grained details which are likely to be overlooked. 
For example, in Figure~\ref{fig:example_jetsking}, the base model only identifies two jet-skiing scenes, while the last one tend to be overlooked due to the surrounding ocean scenes. 
But S2V still captures this scene and thus obtains the correct answer.

\begin{figure*}[t]
  \centering
  \includegraphics[width=1\textwidth]{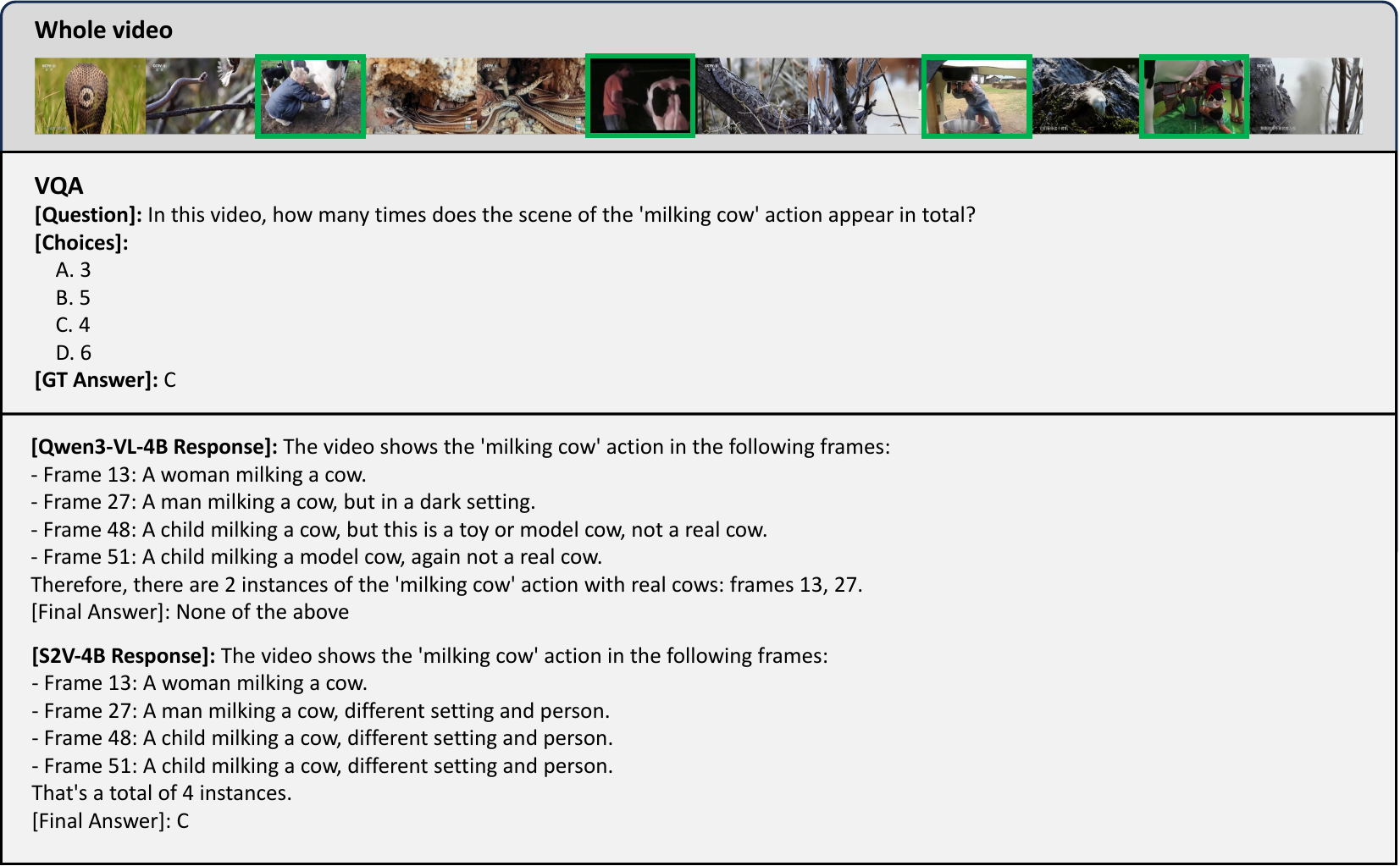}
  \caption{An S2V prediction output successfully associates localized information with correct answer.}
  \label{fig:example_cow}
\end{figure*}

\begin{figure*}[t]
  \centering
  \includegraphics[width=1\textwidth]{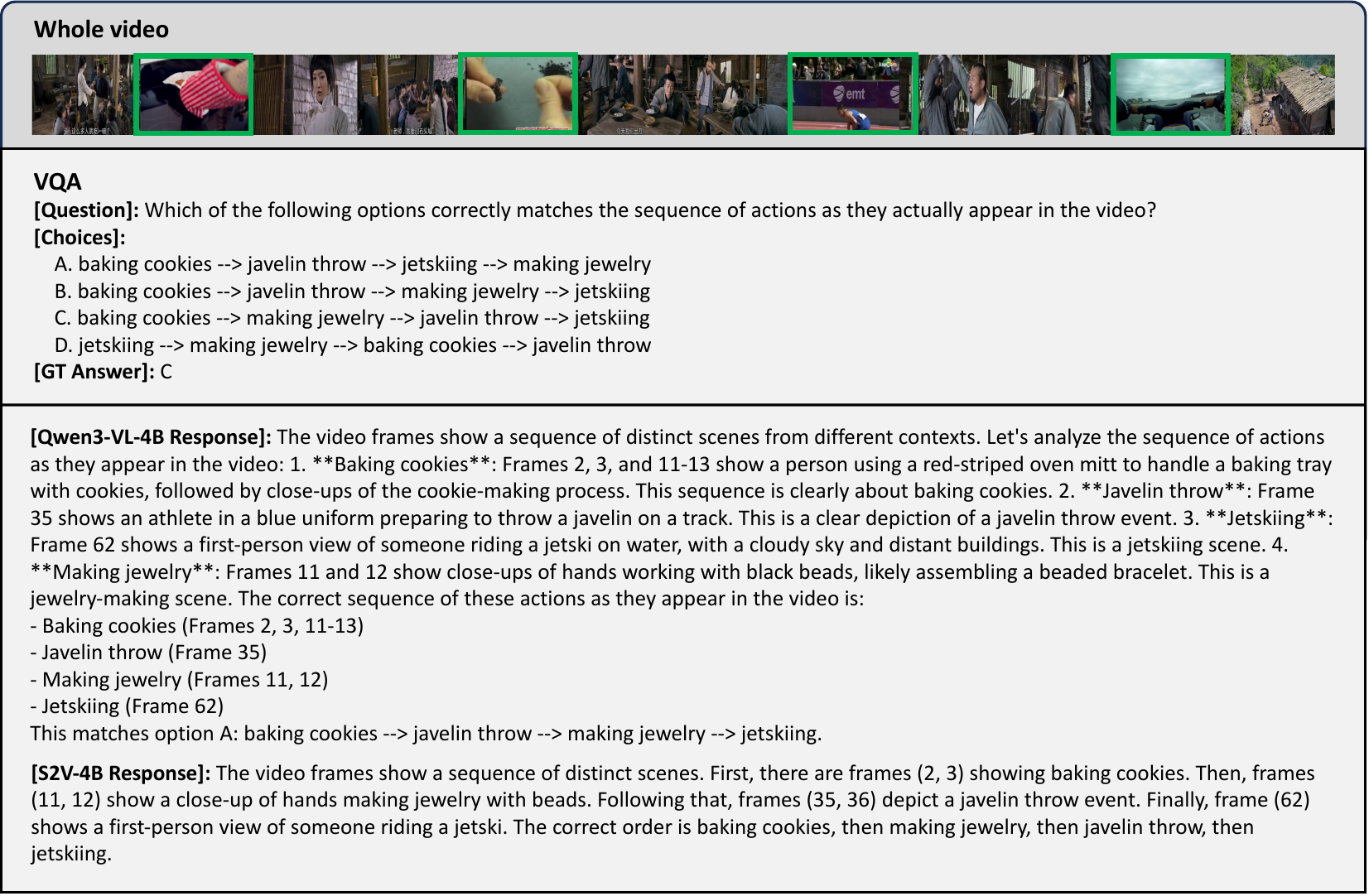}
  \caption{An S2V prediction output correctly comprehends the temporal relations among localized evidence.}
  \label{fig:example_order}
\end{figure*}

Second, S2V strengthens comprehension of fine-grained evidence. 
For example, in Figure~\ref{fig:example_cow}, both the base model and S2V identify four scenes of milking cows. 
However, the base model regards two of them as toy cows and excludes them from counting, leading to an incorrect answer. 
S2V correctly recognizes that all of them correspond to milking cows, thereby successfully associating visual evidence with the correct answer. 
And in Figure~\ref{fig:example_order}, although the base model identifies all query-relevant actions, it misjudges their temporal order and produces an incorrect answer while S2V correctly orders these localized clues.

\FloatBarrier

\end{document}